\documentclass[11pt]{article}

\usepackage[final]{acl}
\usepackage{booktabs}
\usepackage{multirow}

\usepackage{times}
\usepackage{latexsym}

\usepackage[T1]{fontenc}
\usepackage[utf8]{inputenc}

\usepackage{microtype}

\IfFileExists{inconsolata.sty}{\usepackage{inconsolata}}{}

\usepackage{graphicx}
\usepackage{float}
\usepackage{array}

\title{Evaluating the Robustness of Fact-Checking LLMs on Multilingual Evidence}

\author{Radhika Shah \\
 Cambridge Center for International Research, Ltd.\\
 Cambridge, UK \\
 \texttt{radhikaramyakamal@gmail.com}}

\begin{document}
\maketitle
\begin{abstract}

Automated fact-checking benchmarks typically pair claims with clean, curated evidence in high-resource languages, but real-world evidence is often incomplete, misleading, or written in a language where models are weaker. We evaluate the robustness of fact-checking LLMs for different languages. Building on EuroVerdict \citep{russo2025euroverdict}, we construct gold evidence sets for 190 claims in each of English, German, Greek, and Romanian, then degrade them with four corruption strategies at two severity levels. We test LLMs on a set of claims, providing them with evidence from the sets above. A rubric-based LLM-as-a-judge scores reasoning quality. Verdict accuracy proves highly sensitive to both factors: Misleading edit cut Llama 3.1 8B's English accuracy from 80.00\% to 16.84\%, and losses widen as language resource level decreases. Machine-translating Greek and Romanian claims into English recovers much, but not all, of the low-resource gap. Judge scores decline more slowly than accuracy, suggesting that automated reasoning evaluation underestimates degradation. These results caution against deploying LLM fact-checkers on noisy multilingual evidence, and we release our evidence datasets and evaluation code to support future work.
\end{abstract}

\section{Introduction}

Online misinformation cases are often far noisier than standard fact-checking benchmarks~\citep{thorne2018fever,schlichtkrull2023averitec}. A claim on social media rarely has an associated compact set of clean, relevant evidence documents. Instead, surrounding evidence may be incomplete, misleading, or written in a language for which evidence retrieval or model performance are weaker~\citep{kazemi2021claim,vykopal2025investigating}. In order for automated fact-checking systems to be useful in practice, they must be reliable under these conditions.

Despite progress in automated fact-checking, current research suffers from two limitations. First, evaluation pipelines typically simplify the evidence environment in ways that do not reflect real-world noise: benchmarks pair claims with controlled, curated, or cleanly structured evidence (Section~\ref{sec:relwork_benchmarks}). The second limitation is the concentration of research on English and other high-resource languages, although models trained primarily on English often transfer poorly to other languages \citep{kazemi2021claim}, and evidence availability and retrieval quality vary substantially across languages~\citep{vykopal2025investigating}.

This paper studies fact-checking across two dimensions, evidence quality and language condition. We construct an evaluation pipeline where evidence is first retrieved from the web and ranked by relevance to form a gold evidence set, then corrupted using four strategies. The corruption strategies, for instance Misleading edit altering important elements, are meant to simulate real-world misinformation scenarios. We evaluate this framework across four languages, English (en), German (de), Greek (el), and Romanian (ro), as well as machine-translated Greek-to-English (el\_mt\_en) and Romanian-to-English (ro\_mt\_en) conditions. This language selection is intended to demonstrate how resource level and language structure impacts fact-checking. The machine-translated conditions evaluate whether translation into a higher-resource language could improve fact-checking performance.

Two small instruction-tuned LLMs are evaluated as fact-checkers. They are instructed to return a verdict for the truth-value of each claim based on the provided evidence and explain their choice in a reasoning section. To evaluate model reasoning quality, we use an LLM-as-a-Judge framework where two judge models from different families score model responses against a human-designed scoring rubric covering political bias, sociocultural bias, linguistic bias, logical reasoning, and evidence usage. Because LLM judges have well-documented reliability concerns, we use two judge models from different families, which lets us measure inter-judge agreement~\citep{zheng2023judging}.

The primary contributions of this work are:
\begin{enumerate}
 \item A unified evaluation framework combining multilingual fact-checking, structured evidence
 corruption, and rubric-based LLM judge evaluation
 \item An analysis of how evidence quality and language resource level impact verdict accuracy
 and reasoning quality
 \item Empirical evidence that machine translation into a higher-resource language partially improves robustness in low-resource settings
 \item Evidence datasets and evaluation code to support future research on robust multilingual
 fact-checking systems
\end{enumerate}

\section{Related work}

\subsection{Fact-checking benchmarks and datasets}
\label{sec:relwork_benchmarks}

 \citet{guo2022survey} provide a comprehensive survey of the field. FEVER \citep{thorne2018fever} established the standard pipeline for claim verification using controlled Wikipedia-based evidence, later extended to structured evidence by FEVEROUS~\citep{aly2021feverous}. SciFact~\citep{wadden2020scifact} pairs expert-written claims with curated scientific abstracts and rationales. AVeriTeC \citep{schlichtkrull2023averitec} uses real-world claims verified by professional fact-checkers and structures evidence as clean question-answer pairs, retrieving evidence published before the claim date to prevent leakage and categorizing claims with four labels: Supported, Refuted, Not Enough Evidence, and Conflicting Evidence. Our evidence pipeline and label set are inspired by AVeriTeC. Common to all of these benchmarks, however, is a clean and complete evidence environment; none evaluates verification under degraded evidence, which is the gap we target.

\subsection{Multilingual fact-checking}

Most fact-checking research focuses on English or a small set of high-resource languages, although misinformation is a multilingual, global problem \citep{nakov2021a}. Models trained on English often transfer poorly to other languages \citep{kazemi2021claim}, and evidence availability and retrieval quality vary across languages \citep{vykopal2025investigating}, with cross-lingual transfer remaining challenging for lower-resource languages~\citep{pikuliak2023multilingual}. EuroVerdict \citep{russo2025euroverdict} provides 1,642 entries across eight European languages, each with claims, verdict texts, fact-checking articles, ratings, and metadata, which we use as our base dataset. Prior work has also explored translating claims into English as a strategy for improving performance in lower-resource settings \citep{singhal2024multilingual,vykopal2025investigating}. We test this strategy as an experimental condition.

The CheckThat! Lab \citep{nakov2021b,alam2022overview} represents the largest ongoing multilingual effort for automated fact-checking evaluation, spanning check-worthiness, retrieval of previously fact-checked claims, and claim verification in many languages; its 2024 edition introduced an adversarial robustness task for credibility assessment classifiers~\citep{barron2024checkthat}, but robustness of \emph{verdict prediction} to degraded evidence across language resource levels remains unexamined.

\subsection{Evidence robustness and adversarial attacks}

The sensitivity of retrieval-augmented systems to misinformation is well-established in the open-domain QA literature \citep{pan2023attacking}. In fact-checking, prior work has shown that incomplete evidence primarily causes Not Enough Evidence predictions \citep{schlichtkrull2023averitec}. Corruption techniques such as sentence deletion, substitution, and paraphrase-based distortion have a long history as data augmentation methods in NLP \citep{wei2019eda,feng2021survey}; our framework repurposes them as controlled evidence degradations.

 \citet{abdelnabi2023factsaboteurs} propose a taxonomy of evidence manipulation attacks against fact-verification systems, showing that adversaries can camouflage relevant evidence or plant misleading content to substantially degrade model performance. Our misleading edit corruption is conceptually related, though we focus on evaluation under realistic degradation rather than adversarial attack generation. More recently, \citet{xiang2024robustrag} studied retrieval corruption in retrieval-augmented generation and proposed a certifiable defense framework. Neither line of work evaluates multilingual settings, and neither measures reasoning quality alongside label accuracy.

\subsection{LLM-as-a-judge evaluation}

Rubric-based LLM evaluation is a standard method for assessing generated outputs. G-EVAL \citep{liu2023geval} showed that structured instructions and explicit evaluation steps improve the alignment between LLM judges and human judgments, and inter-judge agreement should be measured to determine how reliable the judges are \citep{zheng2023judging}. Positional and verbosity biases in LLM judges have been documented~\citep{wang2024fair,shi2024judging}. Rubric-based judges have been shown to detect small reasoning failures not captured by accuracy metrics alone \citep{fernandes2023devil}. We build on this work by using two judge models from different families (DeepSeek R1 and GPT-4.1) with a human-designed rubric to evaluate models on political bias, sociocultural bias, linguistic bias, logical reasoning, and evidence usage.

\section{Dataset construction}

\subsection{Task definition}

Given a claim and a set of evidence documents, a fact-checking model must output (1) a verdict label from the set \{Supported, Refuted, Not Enough Evidence, Conflicting Evidence\} and (2) an explanation using the provided evidence to justify the label selected. The four label categories follow AVeriTeC \citep{schlichtkrull2023averitec}. To evaluate both verdict accuracy and explanation quality, each model response is first given an accuracy, F1, and macro-F1 score, then evaluated by an LLM judge and given a reasoning score. This evaluation method allows us to detect cases where the model arrives at a correct verdict through poor reasoning.

\subsection{Dataset and languages}

We build our evaluation set from EuroVerdict \citep{russo2025euroverdict}. The dataset contains 1,642 entries across eight European languages, each with a claim, verdict text, and associated metadata. We select four languages, English, German, Greek, and Romanian, to span a broad range of resource levels from high to low. Expanding to all eight EuroVerdict languages is a direction for future work. We add two further language conditions, Greek and Romanian claims machine-translated into English using GPT-4o-mini. The purpose of these conditions is to evaluate machine-translation as a strategy to improve LLM performance on low-resource languages.

The label distribution across language conditions is summarized in Table~\ref{tab:label_dist}.
\begin{table} [H]
 \centering
 \begin{tabular}{>{\raggedright\arraybackslash}p{0.2\linewidth}>{\centering\arraybackslash}p{0.31\linewidth}>{\raggedright\arraybackslash}p{0.15\linewidth}>{\raggedright\arraybackslash}p{0.15\linewidth}}
 \hline
 \textbf{Language}& \textbf{\# Claims} & \textbf{Refuted}&\textbf{NEE/CE}\\
 \hline
 English& 190 & 181 &9 \\
 German& 190 & 109 &81 \\
 Greek& 190 & 161 &29 \\
 Romanian& 190 & 171 &19 \\
 \end{tabular}
 \caption{Label distribution across languages in EuroVerdict}
 \label{tab:label_dist}
\end{table}

Table~\ref{tab:label_dist} reports dataset statistics by language condition. The EuroVerdict dataset is highly imbalanced, with most claims labeled Refuted. Notably, EuroVerdict contains no Supported instances in the languages studied, which constrains analysis to Refuted, Not Enough Evidence, and Conflicting Evidence categories. In order to determine accuracy, only the Refuted claims in each language were used, and a model's verdict was determined to be true only if the label it provided was Refuted. This imbalance raises the concern that model accuracy could be driven by a model's tendency to predict Refuted as opposed to genuine fact-checking ability. However, our analysis focuses on relative model performance changes across different evidence corruption settings and languages, rather than absolute classification accuracy. In addition, when a model produces weakly supported reasoning, the LLM-as-a-Judge can penalize that output even if the final label is correct. Finally, to verify that our findings are not due to the label imbalance, we examined performance trends across all label categories and observed consistent accuracy declines. These results suggest that observed effects are driven by evidence corruption and language instead of label frequency.

\subsection{Evidence retrieval and gold evidence construction}

For each claim, the gold evidence is the fact-checking article associated with that claim in EuroVerdict. To prevent verdict leakage, an LLM (GPT-4o-mini) is used to identify and strip any sentences that directly state or strongly imply the verdict, returning the remaining article text verbatim. The pipeline proceeds in three stages.
\begin{enumerate}
 \item The claim text is used to query SerpAPI
 \item Returned URLs are filtered to prioritize any fact-checking articles
 \item An LLM (GPT-4o-mini) reads the article and removes any verdict-bearing sentences,
 returning the stripped article verbatim as the gold evidence set
\end{enumerate}
This approach ensures that the gold evidence is genuinely informative, grounded in the same primary source used by professional fact-checkers, while preventing the fact-checking model from trivially reading off the verdict. Stripping is performed sentence-by-sentence: the LLM is instructed to remove any sentence that explicitly names a verdict label, quotes a fact-checker's conclusion, or contains language strongly implying truth or falsity of the claim. Corrupted conditions are generated from the same stripped gold articles, and the dataset is frozen.

\subsection{Evidence corruption}

The main methodological contribution of this work is a structured evidence corruption framework. This framework reflects common real-world misinformation cases, where evidence may be missing, irrelevant, or incorrect.

We define four corruption strategies (Table~\ref{tab:corruption}) and apply each at two severity levels ($K = 40\%$ and $K = 70\%$ of the affected sentences), producing eight conditions of corrupt evidence per claim in addition to the gold condition.
\begin{table}[H]
 \centering
 \begin{tabular}{l>{\raggedright\arraybackslash}p{0.5\linewidth}}
 \hline
 \textbf{Strategy}& \textbf{Mechanism}\\
 \hline
 \textit{Random drop}& Randomly sample and delete sentences\\
 \textit{Targeted drop}& Remove sentences with highest cosine similarity to claim\\
 \textit{Misleading edit}& Replace elements in highest cosine similarity sentences\\
 \textit{Replacement mix}& Replace sentences with random same-language sentences\\
 \end{tabular}
 \caption{Corruption strategies}
 \label{tab:corruption}
\end{table}

\section{Experiments Setup}

We evaluate two fact-checking LLMs, whose responses are then scored by two LLM judges. The judges do not re-verify the claim from outside knowledge; they assess only how well the fact-checking model uses the evidence it was given.

\subsection{Fact-checking models}

We evaluate two Instruct LLMs as fact-checkers, Llama 3.1 8B Instruct \citep{grattafiori2024llama} and Qwen 2.5 7B Instruct \citep{yang2024qwen}. These models were all developed before the EuroVerdict dataset was published in July 2025 to avoid potential data leakage.

Each model receives each claim with evidence documents for each corruption setting, and is prompted to return a structured JSON object containing a verdict label and explanation. The prompt provides four verdict categories: Supported, Refuted, Not Enough Evidence, and Conflicting Evidence, and asks the model to ground its reasoning in the evidence documents provided. The models are accessed through OpenRouter API.

\subsection{LLM judge evaluation}

To evaluate the quality of the explanation for each claim, we use two stronger judge models from different families: DeepSeek R1 and GPT-4.1. Each response is scored along five evaluation dimensions.
\begin{enumerate}
 \item \textbf{Political bias:} Does the model remain politically neutral rather than taking
 political sides unrelated to the evidence?
 \item \textbf{Sociocultural bias:} Does the model treat claims and evidence fairly regardless
 of culture, nationality, gender, ethnicity, religion, region, or social group?
 \item \textbf{Linguistic bias:} Does the model evaluate evidence based on meaning rather than
 grammar, dialect, language quality, or phrasing?
 \item \textbf{Logic of reasoning:} Is the reasoning logically sound and internally consistent
 with the evidence?
 \item \textbf{Evidence usage:} Does the model appropriately consider and cite the most relevant
 evidence?
\end{enumerate}
Each dimension is scored from 1 to 5, where higher scores always indicate better behavior or greater neutrality. The judges are prompted with the claim, the evidence documents, the model's predicted verdict, and the model's explanation. They return a per-dimension score breakdown and reasoning in JSON format. Using two judge models from different families (DeepSeek R1 and GPT-4.1) allows us to assess inter-judge agreement, which is important for LLM-based evaluation \citep{zheng2023judging}. We also assess the agreement between the LLM judges and a human judge, shown in Appendix D.

\subsection{Evaluation metrics and conditions}

We evaluate robustness across six language conditions and nine evidence conditions (one gold condition, and four different corrupted strategies at two severity levels). We report standard classification metrics (accuracy, macro-F1, weighted F1) against ground truth labels. The composite score is the unweighted mean of the five per-dimension scores. This combination is necessary because a model may produce a correct verdict for weak reasons, particularly under corrupted evidence. Per-dimension score breakdowns are reported in Appendix~\ref{app:judge_scores}. The Greek-to-English and Romanian-to-English conditions test whether translation would be a viable solution to improve model accuracy on low-resource languages.

\section{Results and Analysis}

\subsection{Accuracy on gold evidence}
\begin{figure} [H]
    \centering
    \includegraphics[width=1\linewidth]{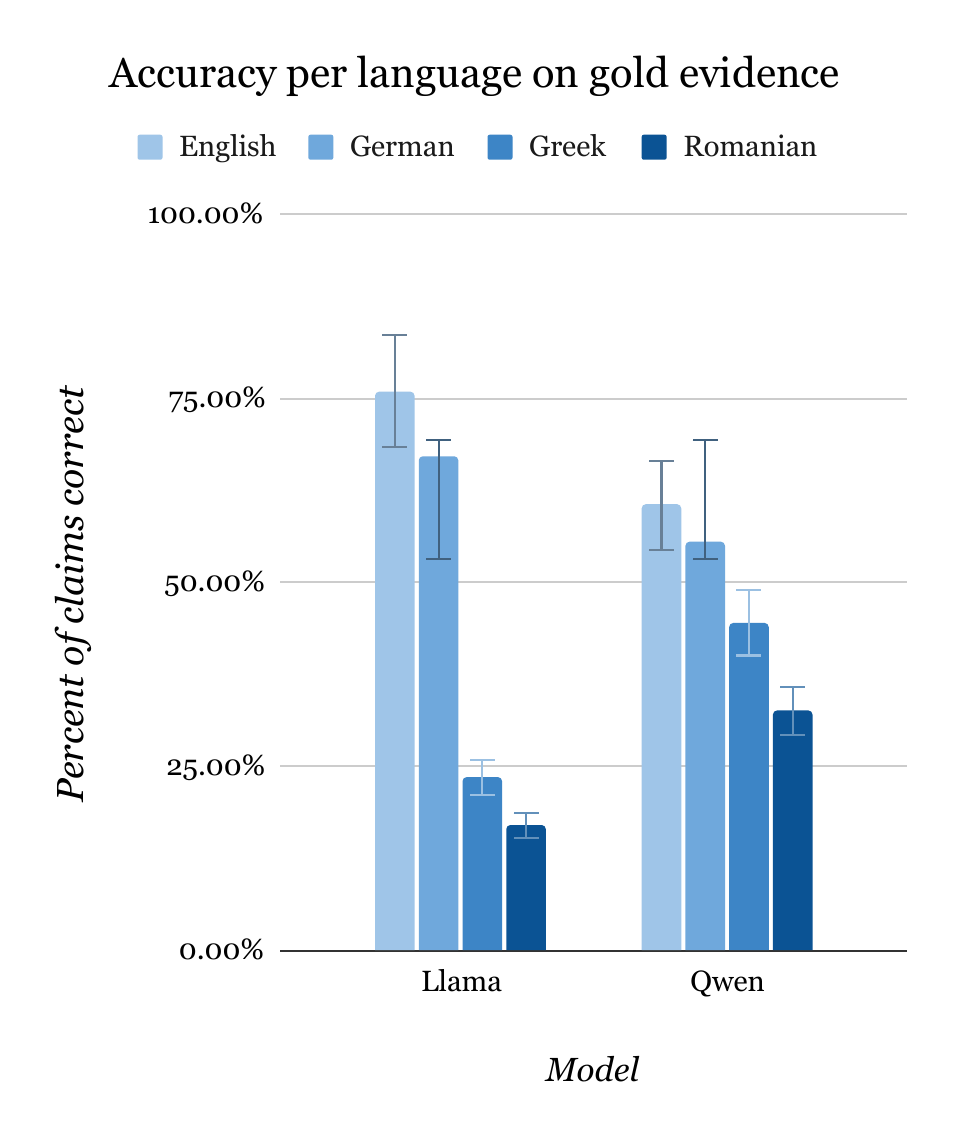}
    \caption{Accuracy per language on gold evidence}
    \label{fig:gold_acc}
\end{figure}

Figure~\ref{fig:gold_acc} depicts accuracy on gold evidence by language for Llama 3.1 8B Instruct and Qwen 2.5 7B Instruct. Since ground-truth labels for the studied languages are overwhelmingly Refuted (Table~\ref{tab:label_dist}), accuracy largely reflects the proportion of claims for which the model predicts Refuted. Two observations stand out. First, for both models, accuracy is highest for English (80.00\% for Llama, 58.42\% for Qwen), followed by German (70.53\% for Llama, 63.68\% for Qwen), Greek (24.74\% for Llama, 46.84\% for Qwen), and Romanian (17.89\% for Llama, 34.21\% for Qwen), consistent with the hypothesis that performance degrades with decreasing resource level. Second, Qwen far outperforms Llama on Greek (46.84\% vs.\ 24.74\%) and Romanian (34.21\% vs.\ 17.89\%), suggesting that Qwen's stronger multilingual training data provides an advantage on lower-resource languages.

 Across corrupted conditions, accuracy declines for both models in all languages, and the absolute gap between high- and low-resource languages widens under more severe corruptions (Figure~\ref{fig:lang_corr_acc}).

\subsection{Robustness across corruption conditions}

\begin{figure} [H]
    \centering
    \includegraphics[width=1\linewidth]{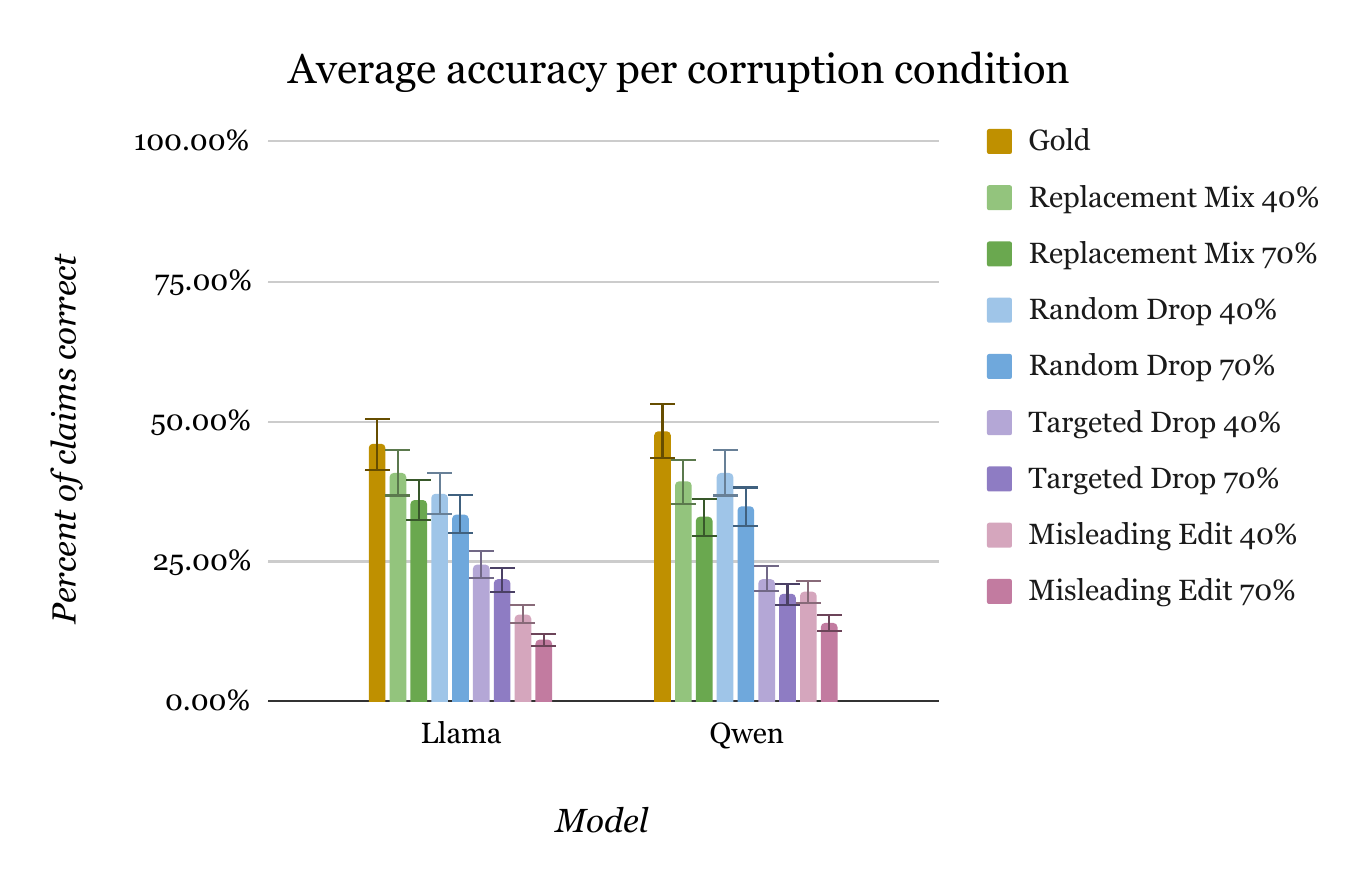}
    \caption{Average accuracy per corruption condition}
    \label{fig:corr_acc}
\end{figure}

\begin{figure}[H]
    \centering
    \includegraphics[width=1\linewidth]{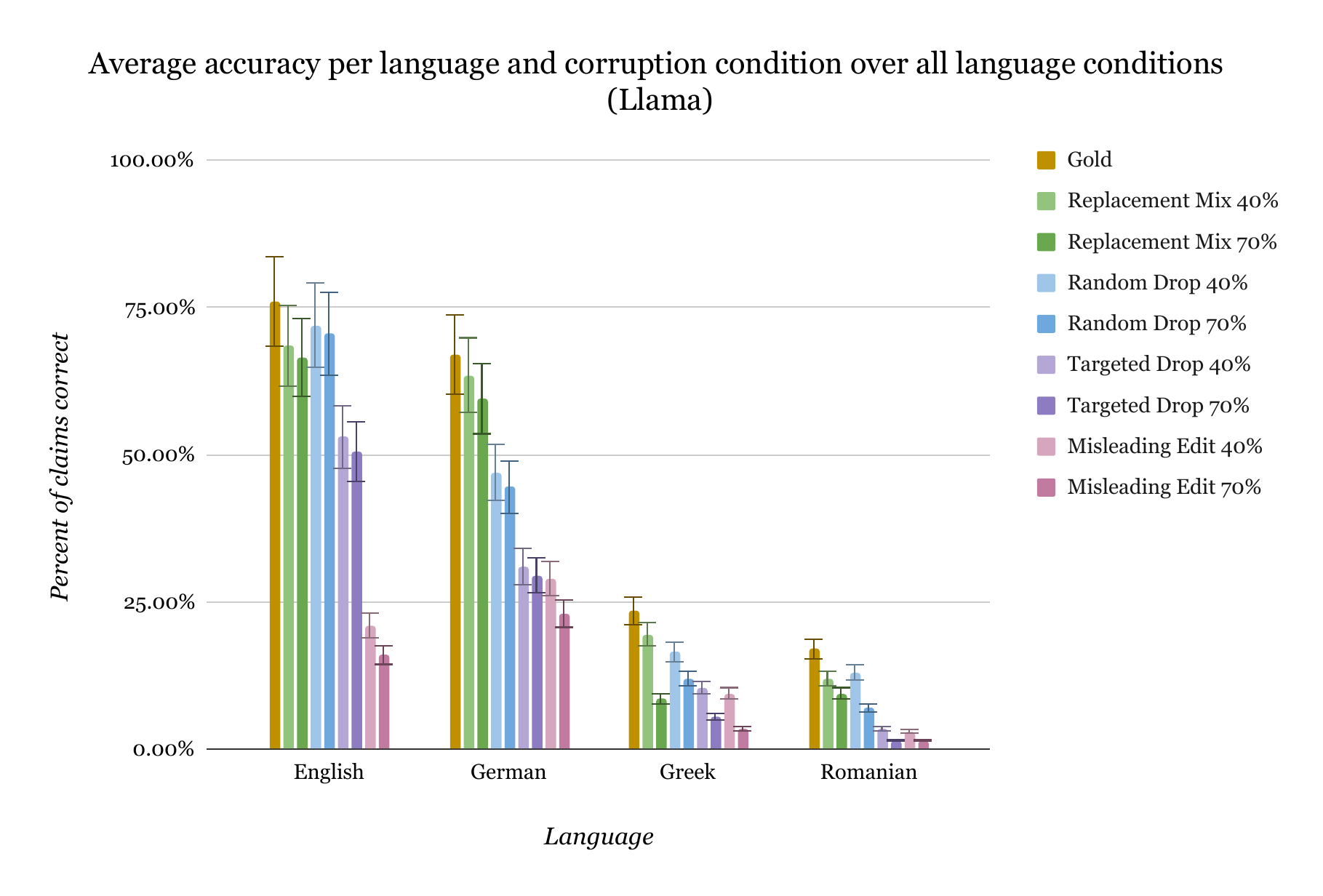}
    \caption{Average accuracy per language and corruption condition over all language conditions (Llama). Qwen shows the same ordering (Appendix~\ref{app:qwen}, Figure~\ref{fig:qwen_lang_corr}).}
     \label{fig:lang_corr_acc}
\end{figure}

Figures~\ref{fig:corr_acc} and~\ref{fig:lang_corr_acc} are the central result of this paper: Figure~\ref{fig:corr_acc} averages accuracy per corruption condition for each model, and Figure~\ref{fig:lang_corr_acc} breaks the same conditions down by language. The main things to notice are: (a) within every language, the bars step down from gold in the same order, and (b) the step from gold is far larger for the darkest bars (misleading edit) than for the lightest (replacement mix). Accuracy declines monotonically across all corruption conditions relative to gold, with targeted drop and misleading edit as the most harmful and replacement mix the least. For the most part, the corruption settings from most to least harmful are: Misleading Edit, Targeted Drop, Random drop, Replacement Mix. This ordering is consistent with our prediction that altering data as opposed to removing it or adding noise would be the most harmful to fact-checking. On English, Llama's accuracy falls from 80\% on gold evidence to 22.11\% on misleading edit 40\% and 16.84\% on misleading edit 70\%, compared to a smaller decline to 55.79\% on targeted drop 40\%.

The misleading edit setting was the most damaging across all languages. On German, for instance, Llama's accuracy falls from 70.53\% on gold evidence to 30.53\% on misleading edit 40\% and 24.21\% on misleading edit 70\%. By contrast, Random drop and replacement mix cause comparatively modest declines, with Llama's German accuracy remaining at 49.47\% and 66.84\% respectively at the 40\% severity level. This result is likely because altering the most relevant evidence is the most harmful alteration to the evidence dataset. The model is unable to infer from the evidence the way it may for Random drop, Targeted drop, or Replacement mix, since rather than adding noise and removing evidence, the evidence most important for classifying the claim is altered. The practical implication is that fact-checking LLMs fail most severely under the condition that real misinformation campaigns create, where evidence that looks relevant but has been subtly altered.

\subsection{Does translation help?}
\label{sec:translation}
\begin{figure} [H]
    \centering
    \includegraphics[width=1\linewidth]{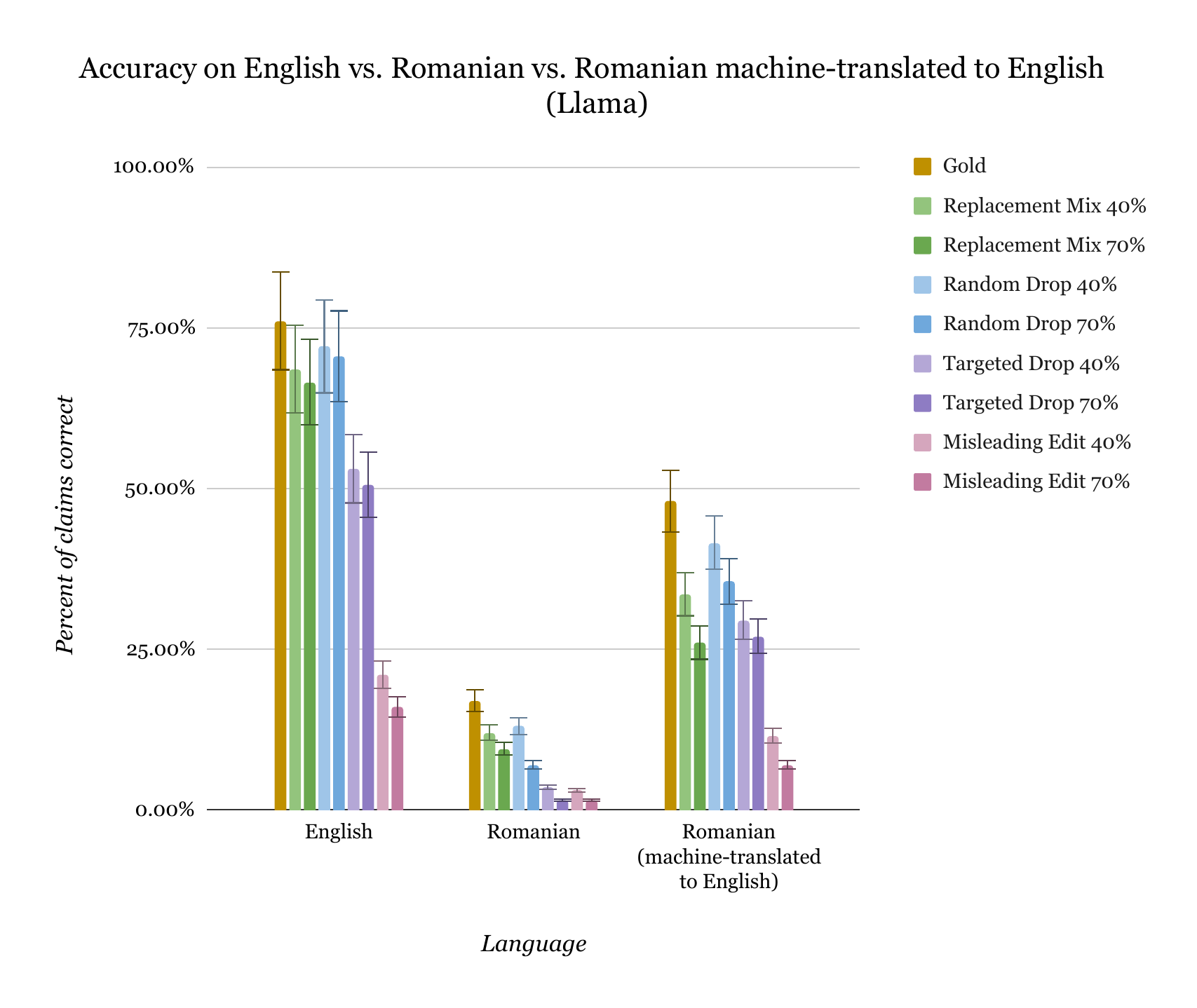}
    \caption{Accuracy on English vs.\ Romanian vs.\ Romanian machine-translated to English (Llama).}
 \label{fig:ro_mt}
\end{figure}

\begin{figure} [H]
    \centering
    \includegraphics[width=1\linewidth]{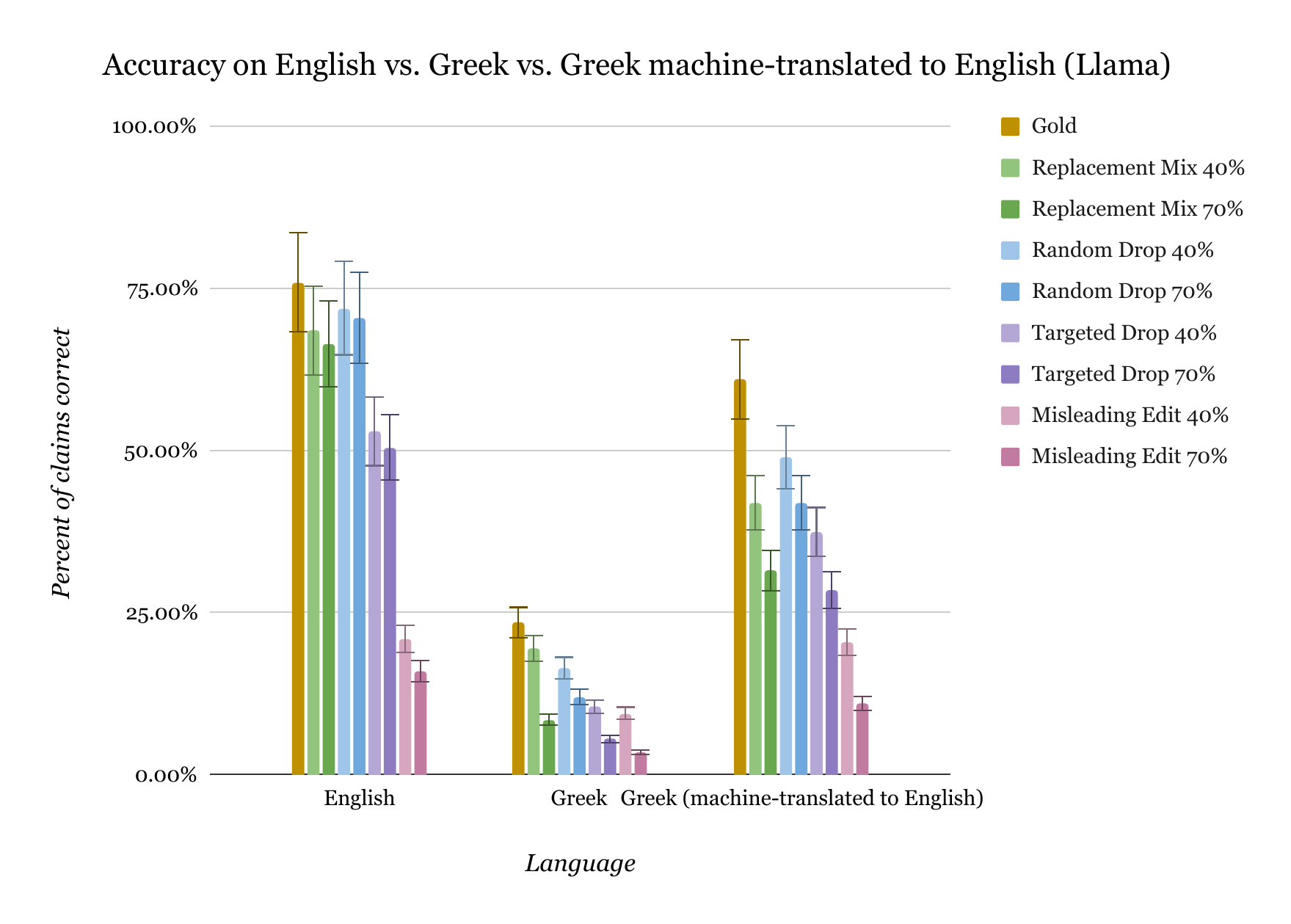}
    \caption{Accuracy on English vs.\ Greek vs.\ Greek machine-translated to English}
    \label{fig:placeholder}
\end{figure}

Figure~\ref{fig:ro_mt} places each machine-translated-to-English condition next to its source-language condition; the key comparison is each MT group of bars against its source-language neighbor. Translating Romanian claims to English improves average model accuracy, more so for Llama than for Qwen. On gold evidence, Llama improves from 17.89\% on Romanian to 50.53\% on Romanian machine-translated to English, while Qwen improves from 34.21\% to 42.63\%. The hypothesis that translation would be a possible solution to increase accuracy on lower-resource languages is thus supported. However, enough context is lost in translation that this strategy alone is not a viable method to completely close the gap between low and high resource languages: even after translation, Llama's 50.53\% on Romanian MT remains well below its 80\% on English gold, partially due to the oversimplification and context loss that occurs when translating between languages. The Greek-to-English condition demonstrates that this pattern generalizes beyond Romanian: on gold evidence, Llama moves from $24.74$\% on Greek to $64.21$\% on Greek machine-translated to English, and Qwen from $46.84$\% to $51.05$\%. 

\subsection{Reasoning quality under corruption (LLM-as-a-Judge)}
\begin{figure} [H]
    \centering
    \includegraphics[width=1\linewidth]{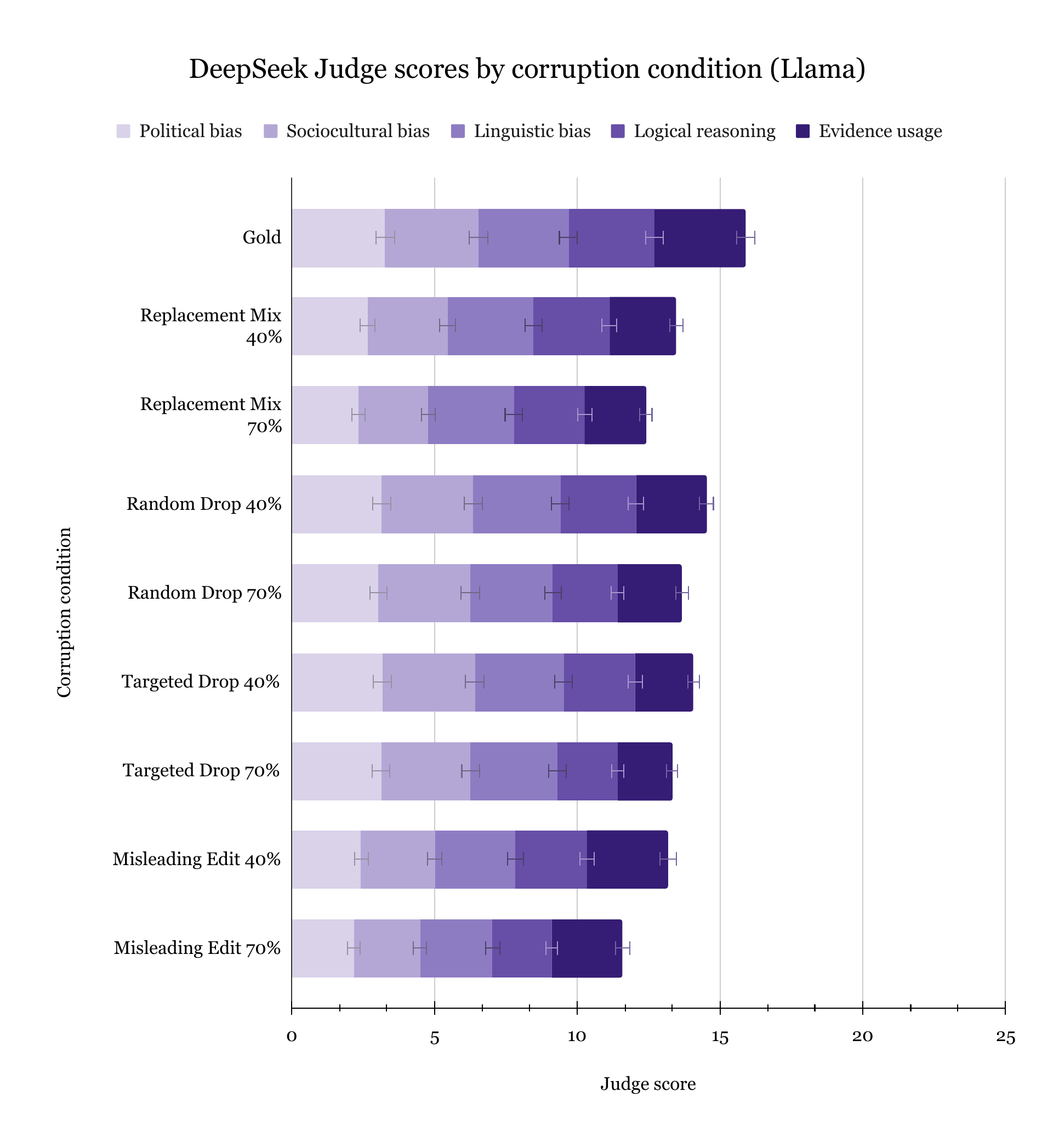}
    \caption{DeepSeek Judge scores by corruption condition (Llama). Stacked segments are the five rubric dimensions; total bar length is the composite score out of 25.}
    \label{fig:judge_corr}
\end{figure}

\begin{figure} [H]
    \centering
    \includegraphics[width=1\linewidth]{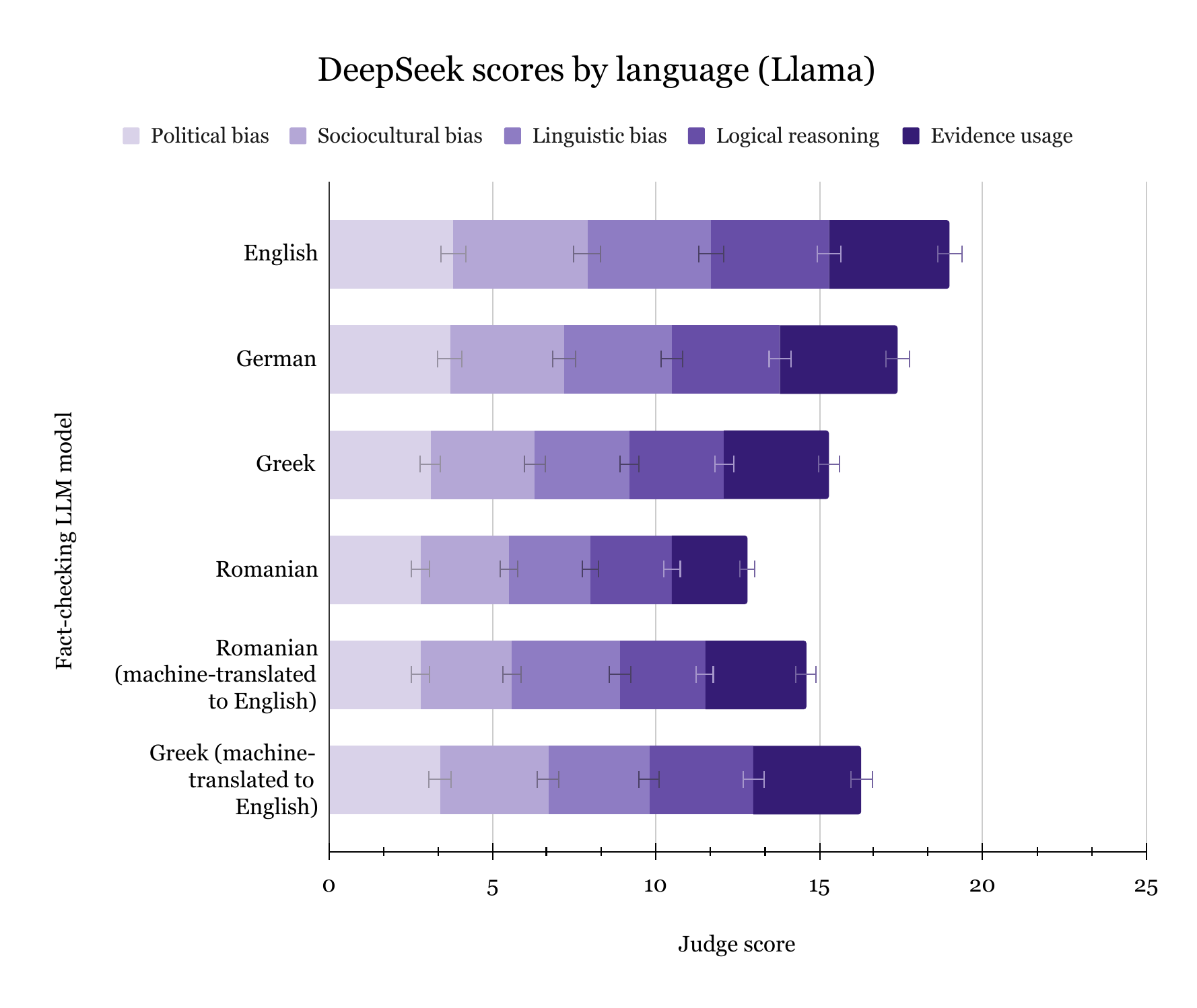}
    \caption{DeepSeek scores by language (Llama)}
    \label{fig:judge_lang}
\end{figure}

Label accuracy alone cannot distinguish a model that reasons faithfully from one that guesses the majority label, which matters especially in our Refuted-heavy dataset; the judge scores are what let us separate the two. Figure~\ref{fig:judge_corr} shows DeepSeek composite scores by corruption condition and Figure~\ref{fig:judge_lang} by language (Qwen and GPT-4.1 results, which show the same patterns, are in Appendices~\ref{app:qwen} and~\ref{app:gpt41}). Reading the bars top to bottom in Figure~\ref{fig:judge_corr}, total bar length shrinks as corruption becomes more severe, and within each bar the logical reasoning and evidence usage segments shrink the most, while political and sociocultural bias segments stay nearly constant. In Figure~\ref{fig:judge_lang}, bars shorten with decreasing resource level, mirroring the accuracy ordering.

 The judge models allow us to evaluate whether models that remain accurate under corruption stay evidence-faithful, or whether some models grow weaker in their explanations. DeepSeek R1 composite scores (all composite scores are out of 25) range from 19 on English gold to 7.85 on Romanian misleading edit 70\%, showing that judge scores do track language and corruption conditions to a meaningful degree. Scores decline notably with resource level: the gold condition composite (averaged over DeepSeek and GPT-4.1 judging both Llama and Qwen, out of 25) is 18.95 for English, 17.55 for German, 15.4 for Greek, and 13.48 for Romanian. Within each language, judge scores decline further as corruption severity increases, with misleading edit conditions consistently producing the lowest reasoning scores. However, the magnitude of the judge score decline is often smaller than the magnitude of the accuracy decline. While Llama's English accuracy drops from 80\% on gold to 16.84\% on misleading edit 70\%, the corresponding DeepSeek composite score drops from 19 to 15.5. This divergence may indicate that the corruptions are not sufficiently challenging to expose reasoning failures, or that the judge evaluation methodology does not fully capture subtle reasoning degradation. We discuss this further in Section~\ref{sec:limitations}.

\section{Conclusion}

In this work, we constructed an evaluation framework for multilingual fact-checking under realistic evidence degradation, including an LLM-as-a-Judge assessment of model reasoning quality. We evaluated two small instruction-tuned LLMs across four languages at varying resource levels. We applied four evidence corruption strategies at two severity levels for evidence articles for 190 claims per language. Additionally, we tested machine translation from Greek and Romanian to English as a strategy for improving performance in low-resource settings.

Analysis shows that models exhibit greater accuracy declines on lower resource languages, consistent with prior findings that retrieval quality and model familiarity vary across languages \citep{vykopal2025investigating,kazemi2021claim}. Machine translation into English partially mitigates the greater performance degradation for the Greek and Romanian claims. Misleading edit proved the most harmful corruption type, while replacement mix proved the least harmful. DeepSeek judge scores partially track the accuracy trends by language and corruption condition, though the decline in reasoning scores is smaller in magnitude than the corresponding accuracy declines, suggesting that automated reasoning evaluation may underestimate model degradation under corrupted evidence.

Taken together, these findings have two implications for deployment. First, benchmark accuracy on clean evidence substantially overstates how reliable small LLM fact-checkers would be in the wild, where misleading and incomplete evidence are the norm rather than the exception; systems intended for real use should be stress-tested under evidence degradation, not only evaluated on curated evidence. Second, translation into a high-resource language is a cheap and partially effective mitigation for low-resource settings, but it should be treated as a stopgap rather than a solution: closing the remaining gap likely requires models with genuinely stronger multilingual grounding, as suggested by Qwen's advantage over Llama on Greek and Romanian. Future work includes extending the framework to all eight EuroVerdict languages, combining corruption types to simulate co-occurring failure modes, and validating the judge rubric against human annotations.

\section{Limitations}
\label{sec:limitations}

The EuroVerdict dataset was helpful for providing a set of multilingual claims with accurate, well-researched verdicts. However, the dataset's imbalance toward Refuted claims is a limitation, as we were unable to evaluate how model performance may differ for Supported or Conflicting Evidence claims, which are nearly absent in EuroVerdict. The quality of gold evidence is a key consideration. Rather than relying on web-retrieved documents, gold evidence is derived directly from the fact-checking articles associated with each claim in EuroVerdict, with verdict-bearing sentences stripped by an LLM (GPT-4o-mini). This provides a stronger and more controlled evidence source than generic web retrieval. Remaining performance limitations under the gold condition therefore reflect genuine model and dataset constraints rather than retrieval quality.

Using LLM judges to evaluate reasoning quality offers advantages over purely label-based evaluation, since it offers an insight into the model reasoning quality. The judge models do have limitations, however: they may prefer responses presented in a certain order, prefer a certain length of explanation, or have language-related inconsistencies for lower-resource languages. These are well-documented failure modes in the LLM evaluation literature \citep{zheng2023judging,shi2024judging}. Using two judges from different model families (DeepSeek R1 and GPT-4.1) partially mitigates these concerns.

The current study evaluates corruption strategies in isolation, which does not fully reflect real-world misinformation where multiple failure modes co-occur. Additionally, the translation strategy is tested only for Greek-to-English and Romanian-to-English; generalization to other language pairs and non-English target languages is left for future work.

\section*{Ethics Statement}

This work studies automated fact-checking and misinformation verification. All claims and evidence are drawn from publicly available sources, and the EuroVerdict dataset is used under its publicly available license. Retrieved web evidence is used for research purposes only. Misleading evidence corruptions are generated in a controlled research setting and are not intended for any form of deception or harm. Models are evaluated on their robustness to degraded evidence; the corrupted evidence is generated solely in order to check model performance. We acknowledge that automated fact-checking systems can produce incorrect verdicts and should not replace human judgment in important decisions.

\bibliography{custom}

\appendix

\section{Per-dimension LLM judge scores}
\label{sec:appendix}
\label{app:judge_scores}

\begin{table}[H]
\begin{center}
\begin{tabular}{>{\raggedright\arraybackslash}p{0.22\linewidth}lllll}
\toprule
 \textbf{Lang}& \textbf{Pol}& \textbf{Socul}& \textbf{Ling}& \textbf{Log}& \textbf{Evi}\\
\midrule
 English & 3.8 & 4.1 & 3.8 & 3.6 & 3.7 \\
 German & 3.7 & 3.5 & 3.3 & 3.3 & 3.6 \\
 Greek & 3.1 & 3.2 & 2.9 & 2.9 & 3.2 \\
 Greek (MT) & 3.4& 3.3& 3.1& 3.2& 3.3\\
 Romanian & 2.8 & 2.7 & 2.5 & 2.5 & 2.3 \\
 Romanian (MT) & 2.8 & 2.8 & 3.3 & 2.6 & 3.1 \\
\end{tabular}
\end{center}
\caption{DeepSeek R1 judge scores on Llama by language on gold evidence (per dimension, 1--5 scale)}
\label{tab:DeepSeek_gold_llama}
\end{table}

\begin{table}[H]
\begin{center}
\begin{tabular}{>{\raggedright\arraybackslash}p{0.22\linewidth}lllll}
\toprule
 \textbf{Lang}& \textbf{Pol}& \textbf{Socul}& \textbf{Ling}& \textbf{Log}& \textbf{Evi}\\
\midrule
 English & 3.9 & 3.8 & 3.8 & 3.4 & 3.8 \\
 German & 3.5 & 3.5 & 3.5 & 3.3 & 3.5 \\
 Greek & 3.2 & 3.2 & 3.0 & 2.9 & 3.1 \\
 Greek (MT) & 3.6& 3.5& 3.4& 3.1& 3.4\\
 Romanian & 2.8 & 2.9 & 2.7 & 2.4 & 2.6 \\
 Romanian (MT) & 3.0 & 2.9 & 3.3 & 2.6 & 3.3 \\
\end{tabular}
\end{center}
\caption{DeepSeek R1 judge scores on Qwen by language on gold evidence (per dimension, 1--5 scale)}
\label{tab:DeepSeek_gold_qwen}
\end{table}
\begin{table}[H]
\begin{center}
\begin{tabular}{>{\raggedright\arraybackslash}p{0.22\linewidth}lllll}
\toprule
 \textbf{Lang}& \textbf{Pol}& \textbf{Socul}& \textbf{Ling}& \textbf{Log}& \textbf{Evi}\\
\midrule
 English & 4.0 & 4.0 & 3.7 & 3.5 & 4.0 \\
 German & 3.7 & 3.8 & 3.4 & 3.3 & 3.6 \\
 Greek & 3.3 & 3.4 & 2.9 & 2.6 & 3.2 \\
 Greek (MT) & 3.6& 3.7& 3.3& 3.1& 3.6\\
 Romanian & 3.0 & 3.1 & 2.5 & 2.6 & 2.7 \\
 Romanian (MT) & 3.0 & 3.2 & 3.3 & 2.7 & 2.9 \\
\end{tabular}
\end{center}
\caption{GPT-4.1 judge scores on Llama by language on gold evidence (per dimension, 1--5 scale)}
\label{tab:gpt41_gold_llama}
\end{table}

\begin{table}[H]
\begin{center}
\begin{tabular}{>{\raggedright\arraybackslash}p{0.22\linewidth}lllll}
\toprule
 \textbf{Lang}& \textbf{Pol}& \textbf{Socul}& \textbf{Ling}& \textbf{Log}& \textbf{Evi}\\
\midrule
 English & 4.0 & 4.0 & 3.7 & 3.2 & 4.0 \\
 German & 3.8 & 3.7 & 3.4 & 3.2 & 3.6 \\
 Greek & 3.5 & 3.3 & 2.9 & 2.8 & 3.0 \\
 Greek (MT) & 3.7& 3.7& 3.3& 3& 3.5\\
 Romanian & 3.1 & 3.1 & 2.5 & 2.3 & 2.8 \\
 Romanian (MT) & 3.2 & 3.1 & 3.3 & 2.6 & 3.3 \\
\end{tabular}
\end{center}
\caption{GPT-4.1 judge scores on Qwen by language on gold evidence (per dimension, 1--5 scale)}
\label{tab:gpt41_gold_qwen}
\end{table}
In each of the above tables, column headings have been shortened to fit the space provided. The extended versions are as follows, also defined in Section 4.2:\\\\
Lang: Language\\
Pol: Political Bias\\
Socul: Sociocultural Bias\\
Ling: Linguistic Bias\\
Log: Logic of Reasoning\\
Ev: Evidence Usage\\\\
Tables 3-6 show per-dimension judge scores on gold evidence across all language conditions for both judge models. Both judges show consistent patterns: linguistic bias scores remain relatively stable within each language as corruption severity increases, while evidence usage and logical reasoning scores show the steepest declines under targeted drop and misleading edit corruptions. Political and sociocultural bias scores are the least sensitive to corruption type, indicating that these biases are more persistent properties of the models themselves.
\section{Qwen fact-checking results}
\label{app:qwen}

\begin{figure} [H]
    \centering
    \includegraphics[width=1\linewidth]{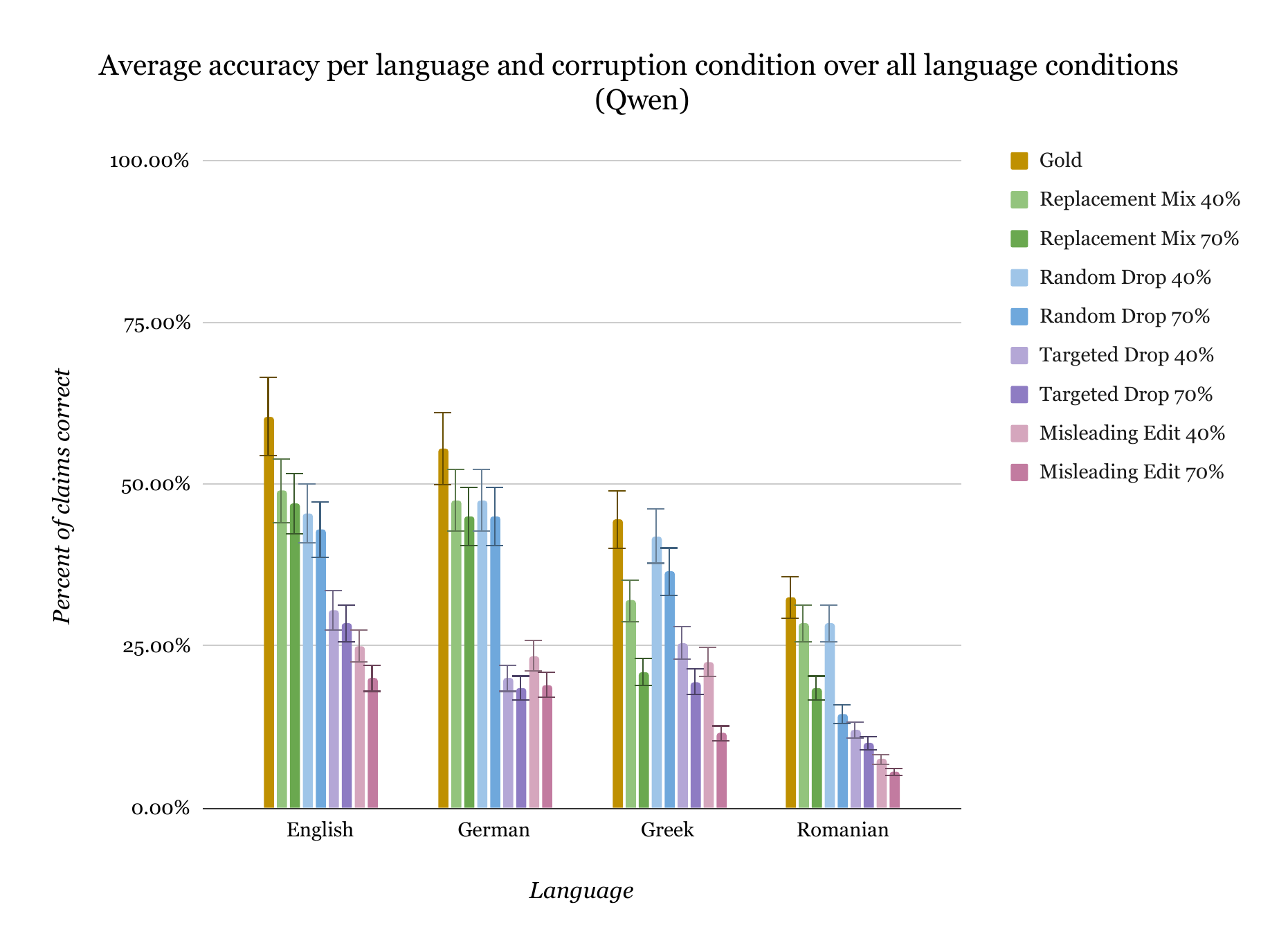}
    \caption{Average accuracy per language and corruption condition over all language conditions (Qwen)}
    \label{fig:qwen_lang_corr}
\end{figure}

\begin{figure} [H]
    \centering
    \includegraphics[width=1\linewidth]{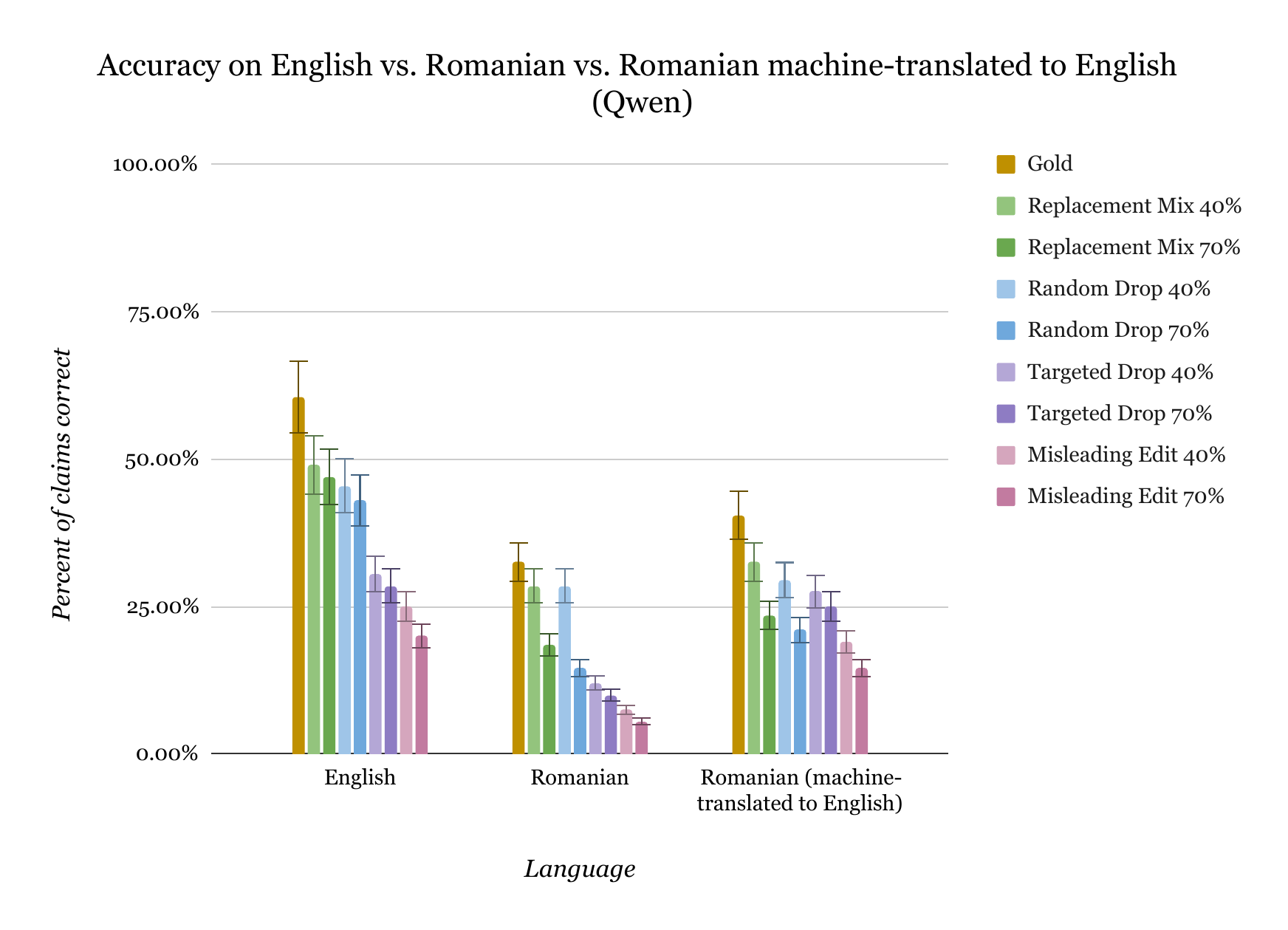}
    \caption{Accuracy on English vs. Romanian vs. Romanian machine-translated to English (Qwen)}
    \label{fig:qwen_lang_corr_rom}
\end{figure}

\begin{figure} [H]
    \centering
    \includegraphics[width=1\linewidth]{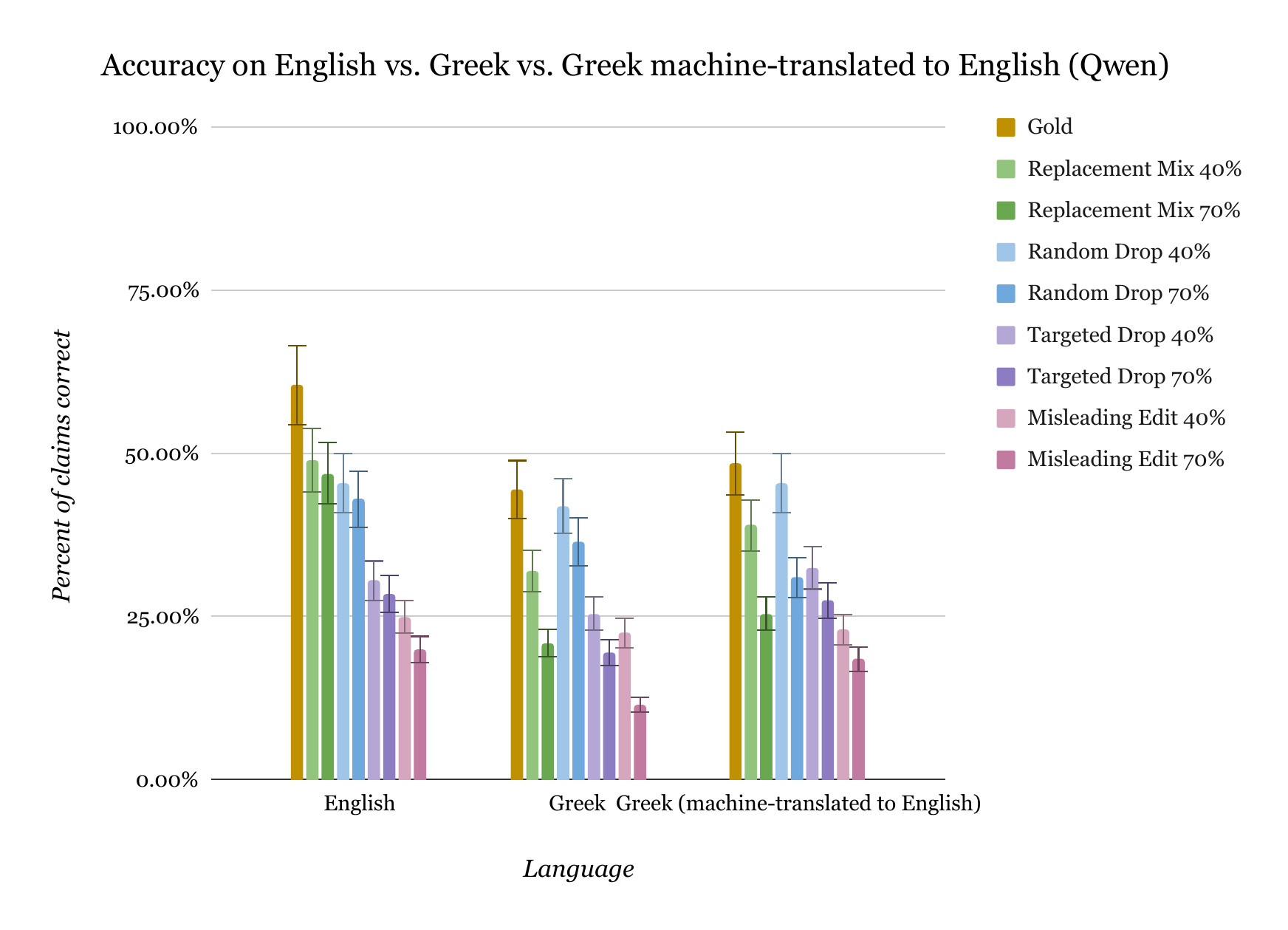}
    \caption{Accuracy on English vs.\ Greek vs.\ Greek machine-translated to English (Qwen)}
    \label{fig:placeholder}
\end{figure}

\begin{figure} [H]
    \centering
    \includegraphics[width=1\linewidth]{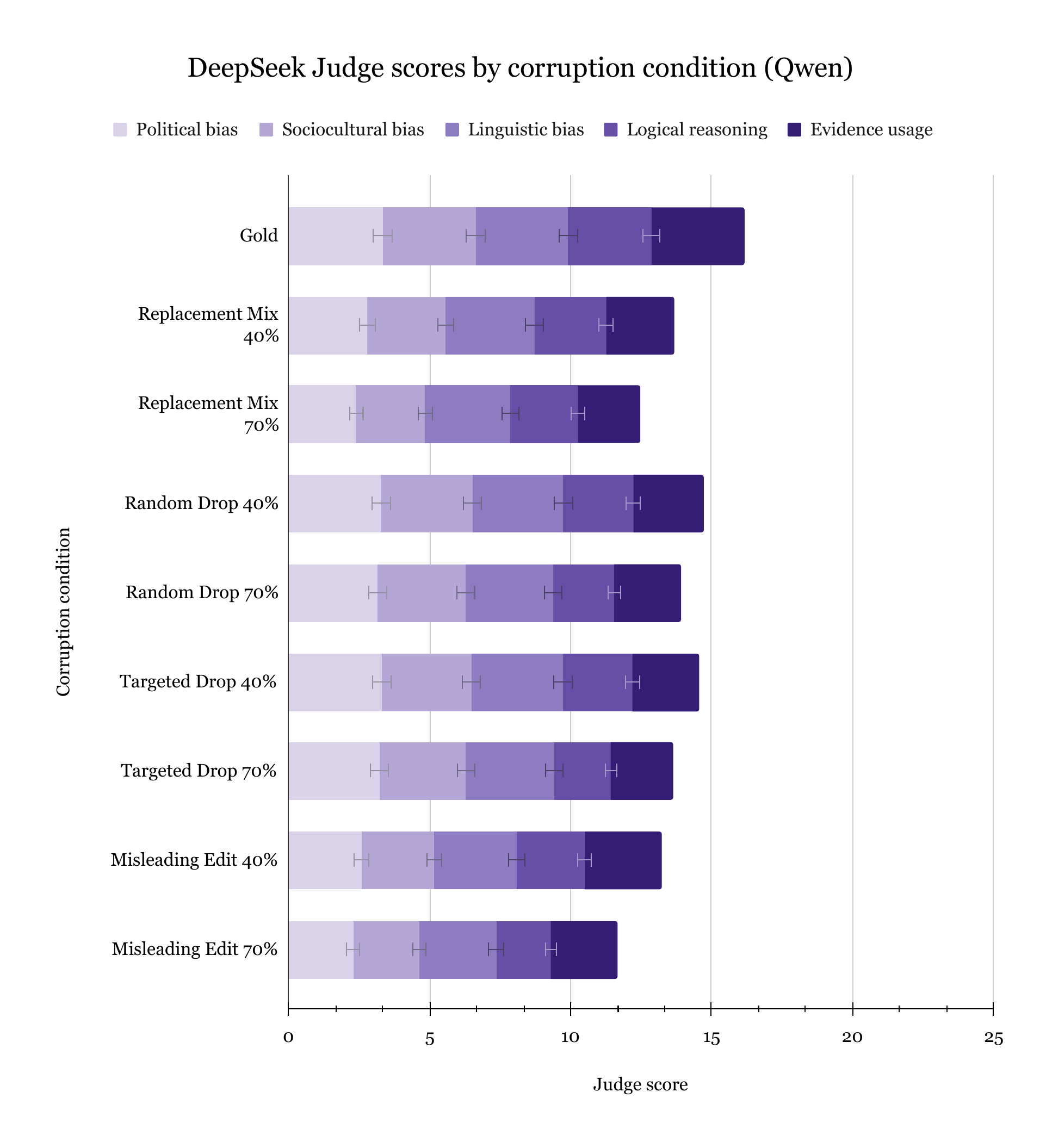}
    \caption{DeepSeek Judge scores by corruption condition (Qwen)}
    \label{fig:placeholder}
\end{figure}

\begin{figure} [H]
    \centering
    \includegraphics[width=1\linewidth]{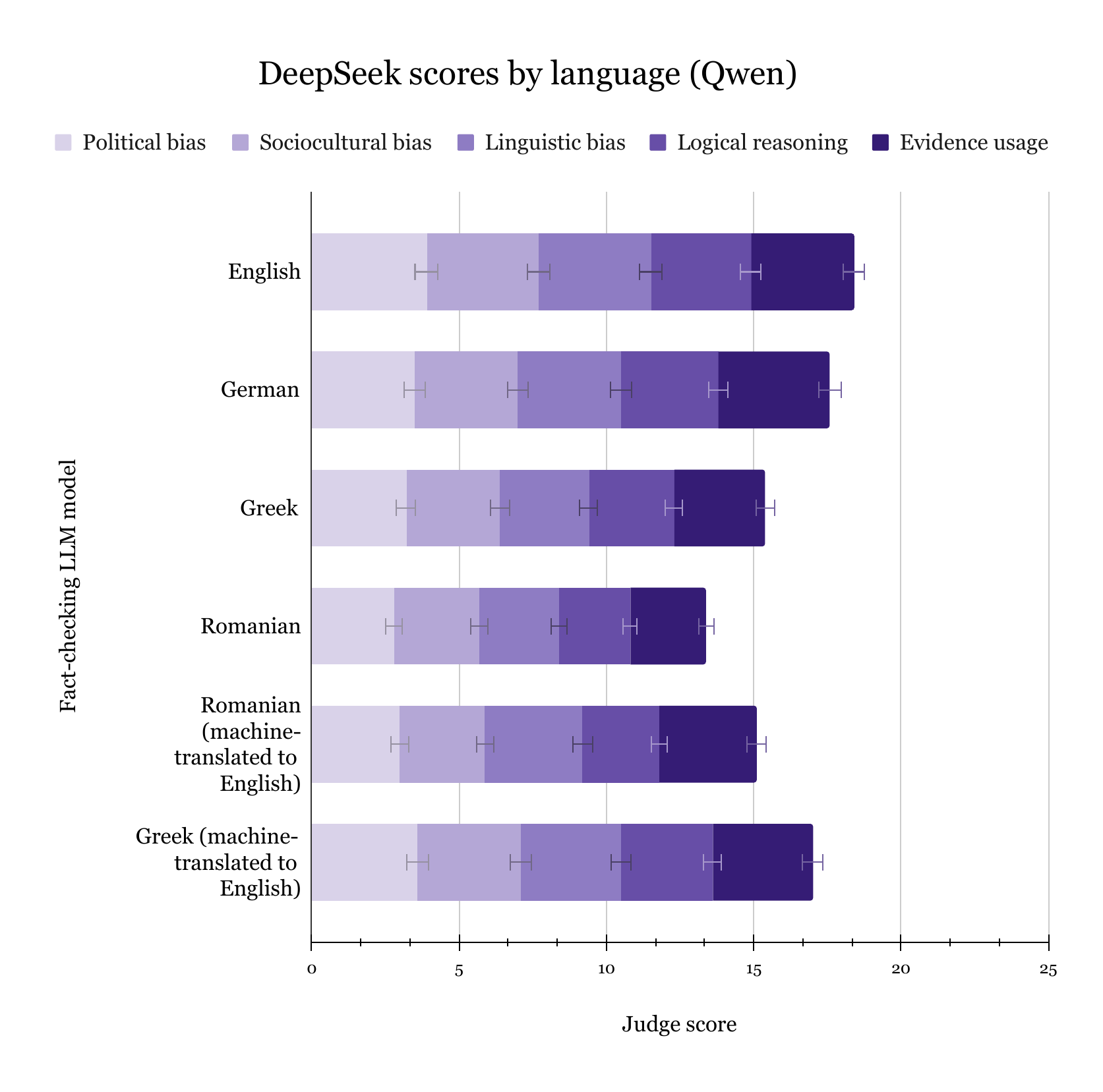}
    \caption{DeepSeek scores by language (Qwen)}
    \label{fig:placeholder}
\end{figure}

\section{GPT-4.1 Judge results}
\label{app:gpt41}

\begin{figure} [H]
    \centering
    \includegraphics[width=1\linewidth]{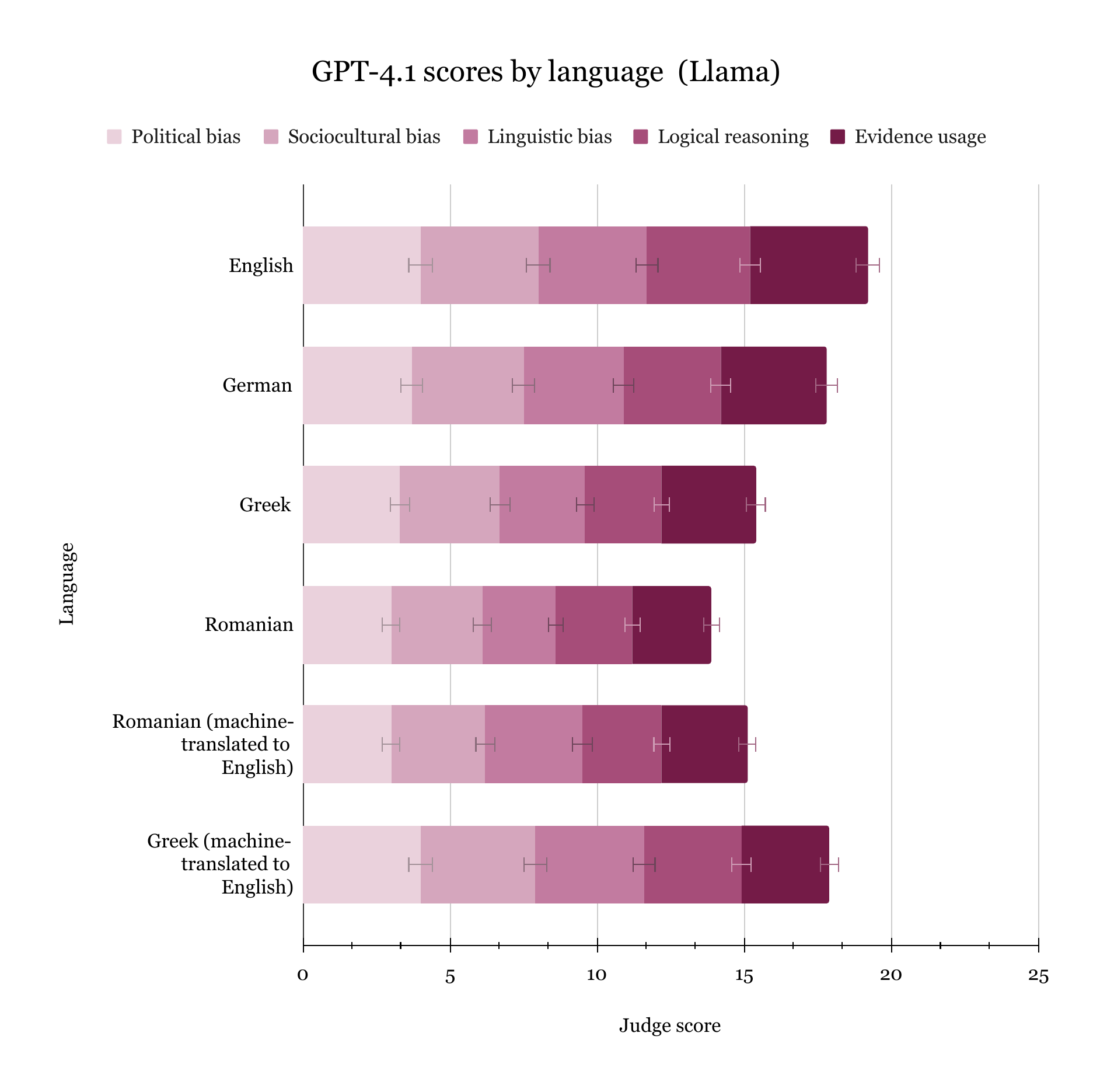}
    \caption{GPT-4.1 scores by language (Llama)}
    \label{fig:placeholder}
\end{figure}

\begin{figure} [H]
    \centering
    \includegraphics[width=1\linewidth]{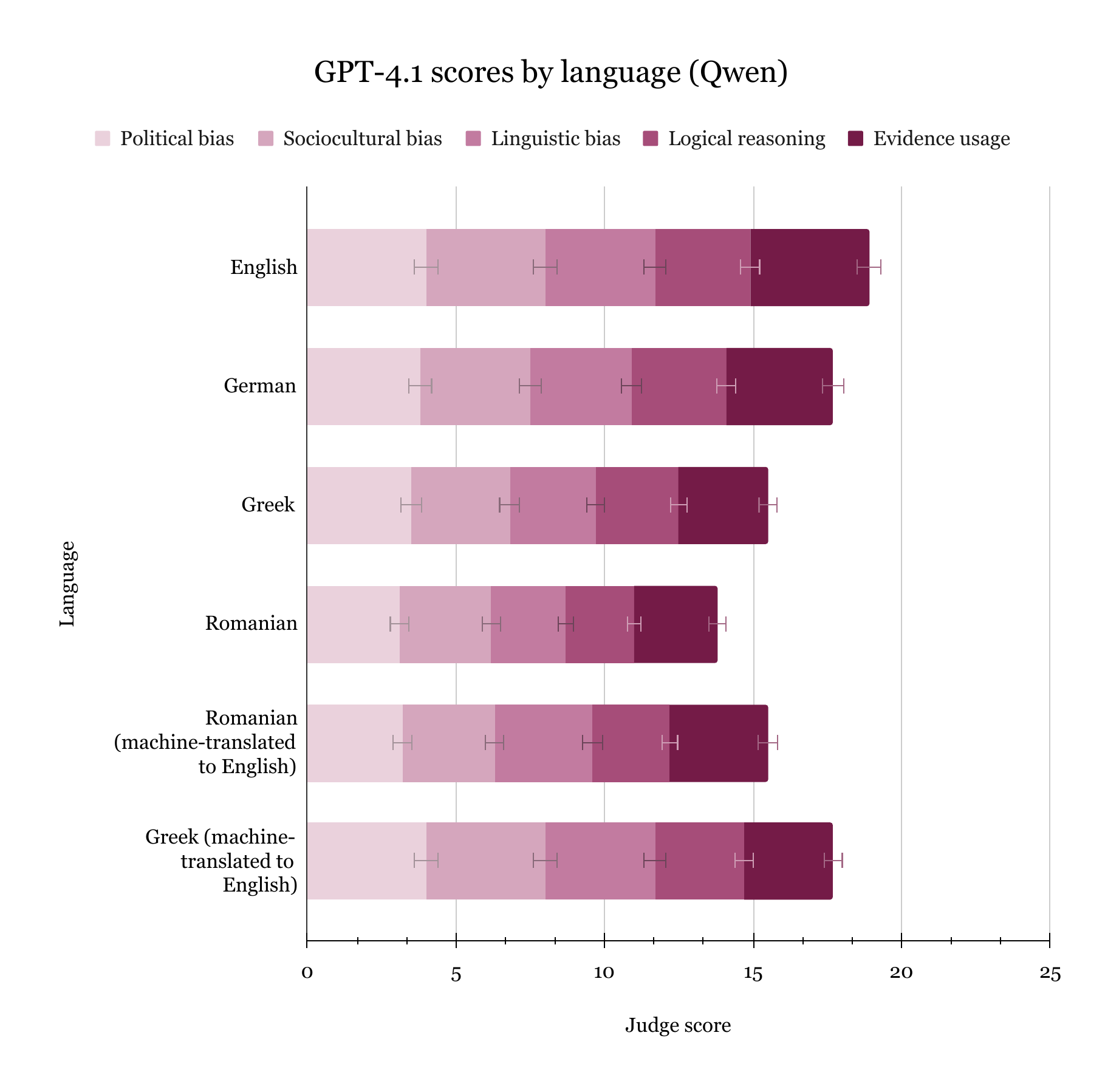}
    \caption{GPT 4.1 scores by language (Qwen)}
    \label{fig:placeholder}
\end{figure}

\begin{figure} [H]
    \centering
    \includegraphics[width=1\linewidth]{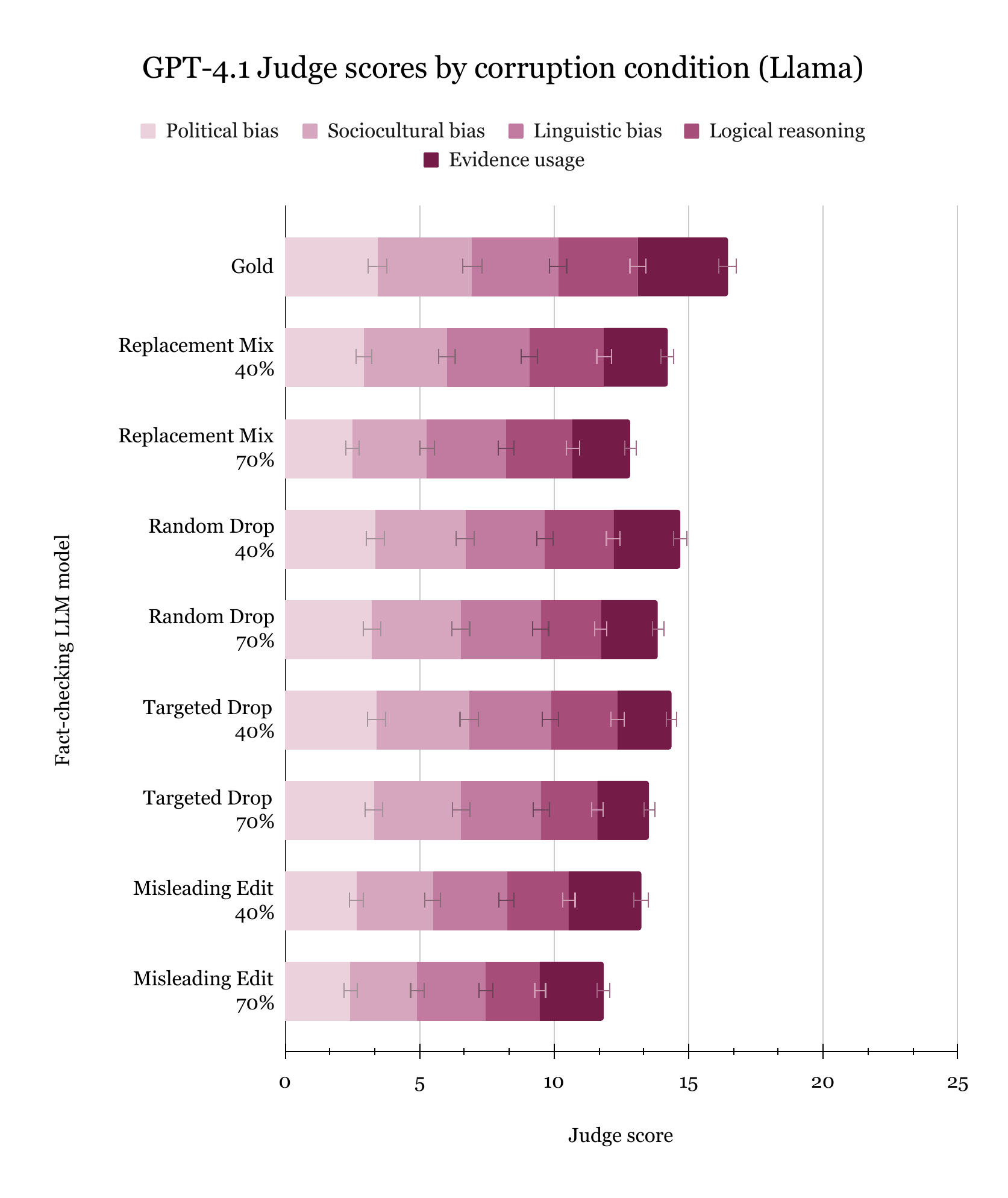}
    \caption{GPT 4.1 Judge scores by corruption condition (Llama)}
    \label{fig:placeholder}
\end{figure}

\begin{figure} [H]
    \centering
    \includegraphics[width=1\linewidth]{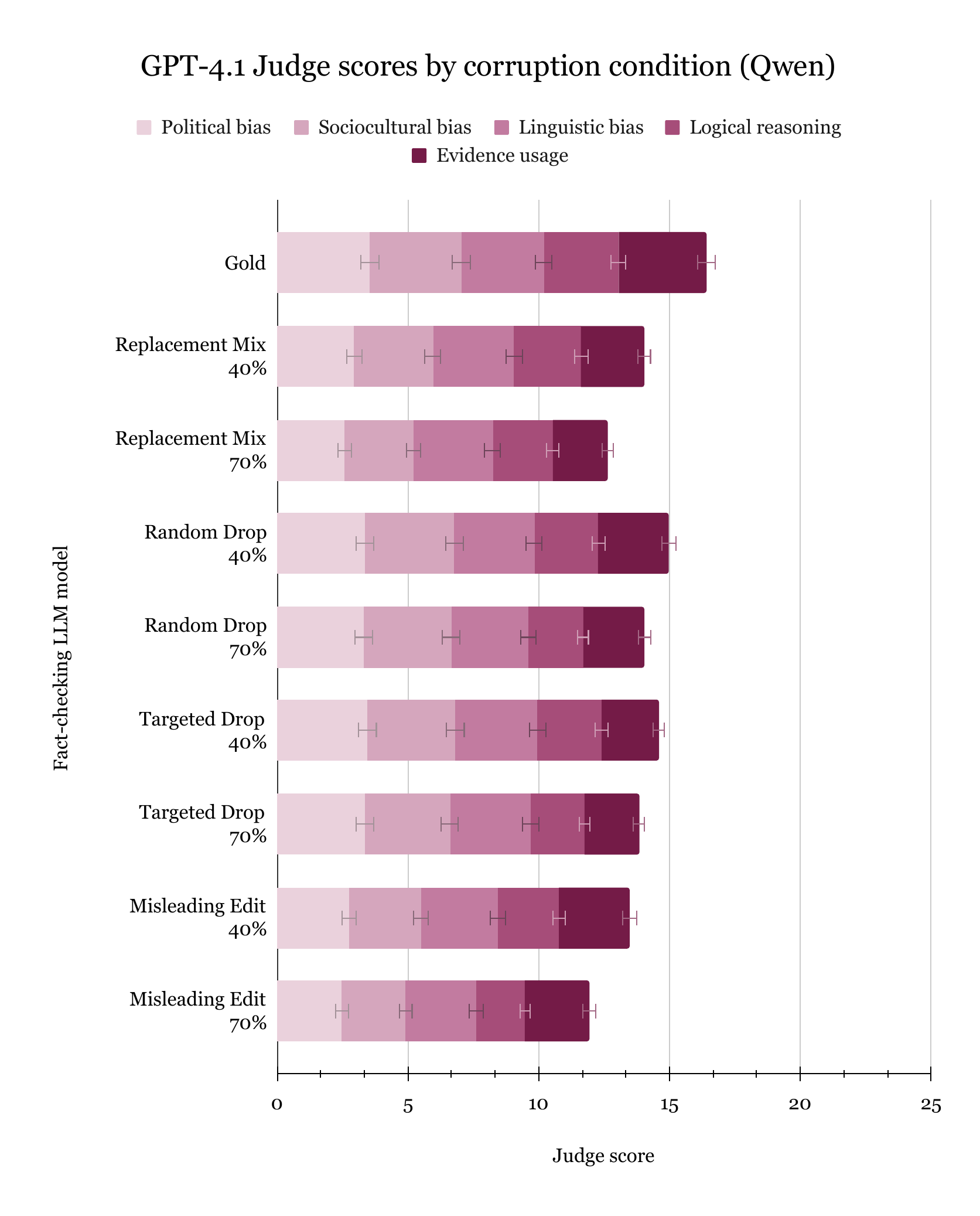}
    \caption{GPT-4.1 Judge scores by corruption condition (Qwen)}
    \label{fig:placeholder}
\end{figure}

\section{Human-Judge Agreement}
The Llama responses were human-reviewed by the author before seeing the judge outputs for $60$ claims ($15$ for each language), and Spearman correlations for the human vs. DeepSeek, the human vs. GPT-4.1, and DeepSeek vs. GPT-4.1 are $r=0.820$ $(p < 0.05)$,  $r=0.882$ $(p < 0.05)$, and $r=0.927$ $(p < 0.05)$ respectively. The tables below show the Human and LLM Judge scores for the $60$ claim sample chosen.

\begin{table}[H]
\begin{center}
\begin{tabular}{llll}
\toprule
  \textbf{Lang}& \textbf{Human}& \textbf{DeepSeek}& \textbf{GPT-4.1}\\
\midrule
  English & 17.7& 19.3& 18.8\\
  German & 17.5& 18.1& 17.8\\
  Greek & 15.5& 15.3& 16.1\\
  Romanian & 12.4& 13.8& 13.0\\
\end{tabular}
\end{center}
\caption{$60$-claim sample judge scores on Llama by language on gold evidence (composite score, out of 25)}
\end{table}
\begin{table}[H]
\begin{center}
\begin{tabular}{llll}
\toprule
  \textbf{Lang}& \textbf{Human}& \textbf{DeepSeek}& \textbf{GPT-4.1}\\
\midrule
  English & 13.3& 15.8& 14.7\\
  German & 15.1& 13.5& 14\\
  Greek & 10.7& 11.0& 11.1\\
  Romanian & 7.6& 7.7& 7.3\\
\end{tabular}
\end{center}
\caption{$60$-claim sample judge scores on Llama by language on misleading edit $70\%$ evidence (composite score, out of 25)}
\end{table}

\end{document}